\documentclass[11pt]{article}

\usepackage[final]{acl}
\usepackage{amsmath, amssymb}  
\usepackage[table]{xcolor}
\usepackage{subcaption}
\usepackage{wrapfig}
\usepackage{multirow}
\usepackage{makecell}

\usepackage{pgfplots}
\pgfplotsset{compat=1.18}
\usepackage{float}
\usepackage{enumitem}
\usepackage{tcolorbox}
\usepackage{listings}
\usepackage{times}
\usepackage{latexsym}

\usepackage[T1]{fontenc}

\usepackage[utf8]{inputenc}

\usepackage{microtype}

\usepackage{inconsolata}

\usepackage{graphicx}

\title{CaptchaMind: Training CAPTCHA Solvers via Reinforcement Learning with Explicit Reasoning Supervision}

\author{
  \textbf{Pengcheng Wang}\textsuperscript{1}
  \textbf{Haoxiang Liu}\textsuperscript{1}
  \textbf{Yang Dai}\textsuperscript{1}
  \textbf{Xiangxiang Zeng}\textsuperscript{1}\\
  \textbf{Guanhua Chen}\textsuperscript{2}
  \textbf{Baotian Hu}\textsuperscript{3}
  \textbf{Longyue Wang}\textsuperscript{1\dag}
  \textbf{Weihua Luo}\textsuperscript{1}
  \\\\
  \textsuperscript{1}Alibaba Group \quad
  \textsuperscript{2}Southern University of Science and Technology\\
  \textsuperscript{3}Harbin Institute of Technology (Shenzhen)\\
  \small{\textsuperscript{\dag}Corresponding author} \\
  \small{\textbf{Code \& Data:} \url{https://github.com/AlibabaResearch/captcha-mind}}
}

\begin{document}
\maketitle

\begin{abstract}
CAPTCHAs are widely deployed as human verification mechanisms and frequently block intelligent agents from completing end-to-end automation in real-world web environments. Solving modern CAPTCHAs requires robust multi-step visual reasoning and interaction capabilities, yet training-based approaches have remained absent due to the lack of large-scale training data and process-level annotations. We introduce CaptchaBench, the first CAPTCHA benchmark designed to support large-scale training, comprising 16,000 programmatically generated samples across eight task categories with detailed region and process-level annotations. Systematic evaluation on CaptchaBench reveals that existing methods fail consistently on tasks requiring fine-grained visual detail capture and region-level comparison. We therefore present CaptchaMind, an RL-based solver trained with explicit reasoning process supervision, achieving 82.9\% average success rate across eight tasks and 71.0\% on real-world instances, substantially outperforming all existing methods without closed-source APIs.

\end{abstract}

\section{Introduction}
Multimodal agents based on large language models (LLMs)~\cite{gpt4,deepseekr1,llama,palm,chinchilla} and vision-language models (VLMs)~\cite{qwen25vl,claude4,gemini3,gpt4o} are rapidly advancing toward real-world deployment, enabling tasks such as web navigation~\cite{mind2web,webvoyager}, UI automation~\cite{aguvis,showui}, and online interaction. However, CAPTCHA (Completely Automated Public Turing test to tell Computers and Humans Apart) remains a critical bottleneck that limits their practical applicability. Modern CAPTCHAs have evolved far beyond early character recognition~\cite{vonahn2003captcha,wang2018captcha}, requiring complex visual reasoning, spatial understanding, and multi-step interaction~\cite{gao2021security,opencaptchaworld}. Recent results on the OpenCaptchaWorld benchmark~\cite{opencaptchaworld} show that even the strongest multimodal agent configurations achieve only around 40\% success, far below human performance, highlighting the significant difficulty of these tasks.

Existing CAPTCHA-solving methods rely on prompting closed-source models~\cite{oedipus,halligan}, incurring high inference costs and lacking controllability. Training-based alternatives appear promising, yet prior work finds that supervised fine-tuning yields poor results~\cite{oedipus}, attributed to the scarcity of training data and high annotation costs. A deeper obstacle is that existing benchmarks such as OpenCaptchaWorld~\cite{opencaptchaworld} are designed exclusively for evaluation --- they provide neither sufficient training data nor the annotations required to supervise intermediate visual reasoning, making it infeasible to develop and validate training-based methods for CAPTCHA solving.

\begin{figure*}[t]
  \centering
  \includegraphics[width=\textwidth]{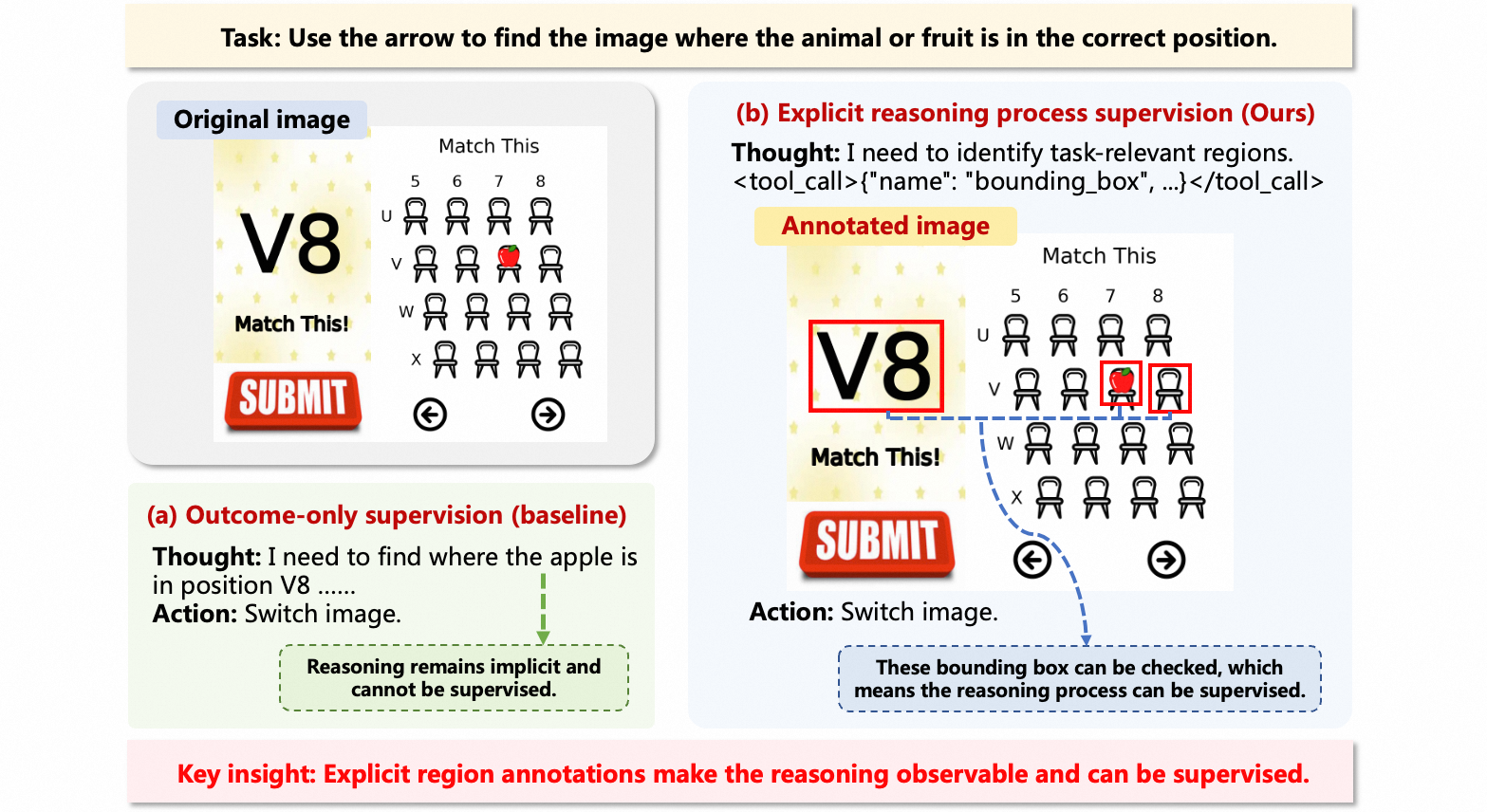}
  \caption{Outcome-only supervision (baseline) vs. explicit reasoning process supervision (ours). In the baseline approach, the model's reasoning process remains implicit and cannot be supervised. Our approach makes intermediate reasoning observable through explicit region annotations via the bounding box tool, enabling direct supervision of which visual elements the model considers task-relevant, beyond final outcomes.}
  \label{fig:overview}
\end{figure*}

To address this gap, we introduce \textbf{CaptchaBench}, the first CAPTCHA benchmark designed to support large-scale training of vision-language agents. CaptchaBench provides 16,000 programmatically generated samples across eight task categories with detailed region and process-level annotations, providing the foundation for analyzing model reasoning behaviors and enabling training-based methods for CAPTCHA solving. 

Systematic evaluation of existing methods on CaptchaBench reveals a consistent pattern: existing methods fail consistently on tasks requiring fine-grained visual detail capture and region-level comparison. Analysis of failure cases suggests that models tend to rely on holistic visual impressions rather than attending to task-relevant details, missing critical region-level information needed for correct decisions. 
Motivated by this finding, we propose \textbf{CaptchaMind}, a training-based solver with explicit reasoning process supervision. As illustrated in Figure~\ref{fig:overview}, unlike outcome-only approaches where the model's reasoning remains implicit and unsupervisable, CaptchaMind makes task-relevant visual regions explicit through bounding-box annotations, enabling direct supervision of which visual elements the model attends to at each reasoning step. Our contributions are as follows:
\begin{itemize}
    \item We introduce CaptchaBench, the first CAPTCHA benchmark designed to support large-scale training of vision-language agents, comprising 16,000 programmatically generated samples across eight task categories with detailed region and process-level annotations. A human discrimination study confirms that our synthesized CAPTCHAs are perceptually indistinguishable from real-world instances.

    \item We conduct systematic evaluation of existing methods on CaptchaBench, revealing that models fail consistently on tasks requiring fine-grained visual detail capture and region-level comparison.

    \item We present CaptchaMind, the first training-based CAPTCHA solver built upon an RL framework with explicit reasoning process supervision that directly rewards correct identification of task-relevant visual regions at intermediate steps.

    \item CaptchaMind achieves state-of-the-art performance across eight tasks, strong generalization to unseen tasks and cross-dataset settings, and 71.0\% success rate on real-world instances without requiring closed-source APIs.
\end{itemize}

\section{Related Work}

\paragraph{CAPTCHA Benchmarks and Solvers.}
Early CAPTCHA research focused on text recognition~\cite{vonahn2003captcha,wang2018captcha}, but modern CAPTCHAs have evolved into complex visual reasoning challenges. OpenCaptchaWorld~\cite{opencaptchaworld} provides a comprehensive evaluation benchmark, revealing that even the strongest models achieve far below human-level performance. Existing solvers such as Oedipus~\cite{oedipus} and Halligan~\cite{halligan} rely on prompting closed-source models, incurring high costs and limited controllability. Oedipus further finds that SFT on small-scale data yields limited improvements. Unlike prior benchmarks designed exclusively for evaluation, CaptchaBench is the first to support large-scale training with process-level annotations, enabling a new class of training-based solvers.

\paragraph{Reasoning Supervision and Multimodal Interactive Training.}
Our work relates to two lines of research: the ``thinking with images'' paradigm~\cite{deepeyes,chainoffocus,openthinkimg,thyme,groundr1,activeo3,minio3}, which enables models to invoke visual tools during reasoning, and GUI/web agents~\cite{uir1,guir1,infiguir1,uitars,seeclick},which combine SFT and RL to learn multi-step action policies through interface interaction. However, in both lines, tool invocations and actions serve primarily as perception aids or control signals, without directly supervising whether intermediate reasoning is correct. CaptchaMind addresses this gap by explicitly rewarding correct region-level grounding at intermediate steps.

\section{CaptchaBench}

\subsection{Tasks and Environment}

\paragraph{Task formulation.}
We formalize the CAPTCHA-solving process as a partially observable Markov decision process (POMDP)~\cite{pomdp}. At time step $t$, the agent receives an observation $h_t$, comprising the current image, task prompt, and interaction history. Based on the policy $\pi_\theta(a_t \mid h_t)$, the agent produces an action $a_t$. The action space includes clicking (click), dragging (drag), typing text (type), and drawing bounding boxes (bounding\_box). The environment maps the action to the next image and interface state according to predefined task logic, resulting in a state transition $s_{t+1} \sim T(s_{t+1} \mid s_t, a_t)$. The environment provides a terminal binary reward of 1 for success and 0 otherwise.

\paragraph{Task types.}
\label{sec:tasks}

We implement eight CAPTCHA tasks drawn from real-world commercial CAPTCHA services (e.g., GeeTest, hCaptcha) and existing benchmarks, grouped into three categories.
 (1) \emph{Image-switching tasks}, including connect\_icon, coordinates, dart\_count, and rotation\_match,where the model switches among candidate images to identify the correct one. (2) \emph{Multi-step interaction tasks}, including click\_order and image\_select, where the model performs multiple interactions on the same image and receives step-wise feedback. (3) \emph{Single-step decision tasks}, including dice\_count and slide\_puzzle, where the model must produce an accurate counting result or a precise dragging action in a single operation. Detailed task interfaces are shown in Appendix~\ref{sec:task_examples}.

\paragraph{Interactive environment.}

We implement a Python-based simulator that provides controllable state transitions, reproducible feedback, and the ability to precisely record key regions and intermediate states. This is critical for supervising and analyzing the model's reasoning behaviors, and provides the foundation for process-level supervision during training.

\subsection{Automated Data Generation}

\begin{table}[t]
\centering
\small
\begin{tabular}{lcc}
\hline
\textbf{Task Type} & \textbf{Train} & \textbf{Test} \\
\hline
connect\_icon & 2,000 & 200 \\
coordinates & 2,000 & 200 \\
dart\_count & 2,000 & 200 \\
dice\_count & 2,000 & 200 \\
rotation\_match & 2,000 & 200 \\
image\_select & 2,000 & 200 \\
click\_order & 2,000 & 200 \\
slide\_puzzle & 2,000 & 200 \\
\hline
\textbf{Total} & \textbf{16,000} & \textbf{1,600} \\
\hline
\end{tabular}
\caption{Dataset statistics for each CAPTCHA task type. Our automated generation pipeline produces 2,000 training samples and 200 test samples per task, yielding a dataset nearly two orders of magnitude larger than OpenCaptchaWorld~\cite{opencaptchaworld}.}
\label{tab:dataset_stats}
\end{table}


To support supervised fine-tuning and reinforcement learning, we design a fully automated data generation pipeline that produces large-scale, diverse training samples with complete process-level supervision signals. Task designs are grounded in real-world commercial CAPTCHA services and existing benchmarks, ensuring that our synthesized instances reflect genuine challenge structures. The pipeline covers programmatic scene construction, multi-step interaction trajectory generation, and chain-of-thought reasoning data generation for certain task types. This fully automated process requires no human annotation and generates 2,000 training samples and 200 test samples per task. Detailed implementation is provided in Appendix~\ref{sec:data_generation}. Table~\ref{tab:dataset_stats} summarizes the dataset statistics.

\subsection{Visual Authenticity Validation}

\label{sec:visual_authenticity}

CaptchaBench's ecological validity rests on two foundations: task designs grounded in real-world commercial CAPTCHA services (Section~\ref{sec:tasks}), and synthesized instances visually indistinguishable from real-world screenshots. To validate the latter, we conduct a human discrimination study.

\paragraph{Study design.}
We collect 100 real CAPTCHA screenshots from live websites and pair them with 100 synthesized samples, covering all eight task types. Fifty volunteers participate, each presented with 100 shuffled images and asked to classify each as Real CAPTCHA or Synthetic CAPTCHA. The experimental interface is shown in Appendix~\ref{sec:study_interface}.

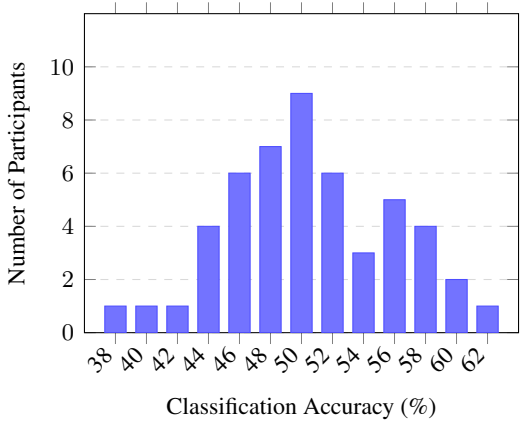
\begin{figure}[t]
\centering
\begin{tikzpicture}

\begin{axis}[
    width=0.95\columnwidth,
    height=5.8cm,
    ybar,
    bar width=8pt,
    xlabel={Classification Accuracy (\%)},
    ylabel={Number of Participants},
    xmin=36, xmax=64,
    ymin=0, ymax=12,
    xtick={38,40,42,44,46,48,50,52,54,56,58,60,62},
    xticklabels={38,40,42,44,46,48,50,52,54,56,58,60,62},
    x tick label style={font=\small, rotate=45, anchor=east},
    ytick={0,2,4,6,8,10},
    ymajorgrids=true,
    grid style={dashed, gray!30},
    tick label style={font=\small},
    label style={font=\small},
]

\addplot[
    fill=blue!55,
    draw=blue!70,
] coordinates {
    (38,1)
    (40,1)
    (42,1)
    (44,4)
    (46,6)
    (48,7)
    (50,9)
    (52,6)
    (54,3)
    (56,5)
    (58,4)
    (60,2)
    (62,1)
};

\end{axis}
\end{tikzpicture}
\caption{Human indistinguishability study results ($N=50$). Bars show the distribution of per-participant classification accuracy (bin width = 2\%).}
\label{fig:human-study}
\end{figure}

\paragraph{Results.}
The mean classification accuracy is 51.04\% ($SD = 5.38\%$), not significantly different from chance ($t(49) = 1.37$, $p = 0.178$). Cohen's Kappa of $\kappa = 0.02$ confirms near-zero agreement with ground-truth labels. The distribution of per-participant accuracy is shown in Figure~\ref{fig:human-study}.

\paragraph{Discussion.}
These results confirm that our synthesized CAPTCHAs are perceptually indistinguishable from real-world instances, validating CaptchaBench as an ecologically valid benchmark. The sim-to-real results in Section~\ref{sec:sim_to_real} further corroborate this finding from a model performance perspective.

\section{CaptchaMind: RL-Based Solver}
\label{sec:baseline_solver}

Our systematic evaluation reveals that existing methods fail consistently on tasks requiring fine-grained visual detail capture, lacking reliable mechanisms to attend to task-relevant regions at intermediate reasoning steps. 
We therefore develop \textbf{CaptchaMind}, an RL-based solver with explicit reasoning process supervision. Our training pipeline consists of two stages: supervised fine-tuning (SFT) to establish basic task-solving capabilities, followed by reinforcement learning with explicit reasoning process supervision.

\subsection{Supervised Fine-Tuning}
\label{sec:sft}
Before reinforcement learning, we perform supervised fine-tuning (SFT) via behavior cloning~\cite{behaviorcloning} to establish basic task-solving capabilities, including correct tool usage (e.g., invoking click, drag, and bounding\_box) and multi-step interaction behaviors. We incorporate chain-of-thought (CoT) reasoning data for representative tasks (dice\_count and rotation\_match) during SFT, which is sufficient for the model to acquire the think-then-act~\cite{cot,react} paradigm across all tasks. 

\subsection{RL with Reasoning Supervision}
\label{sec:rl_supervision}

Building upon the SFT initialization, we further optimize the policy using reinforcement learning. Unlike conventional RL approaches that rely solely on final-outcome rewards, our framework directly supervises the model's reasoning process, improving its ability to capture critical intermediate signals in CAPTCHA tasks.
\paragraph{Making the reasoning process explicit.}

\begin{figure*}[t]
  \centering
  \includegraphics[width=\textwidth]{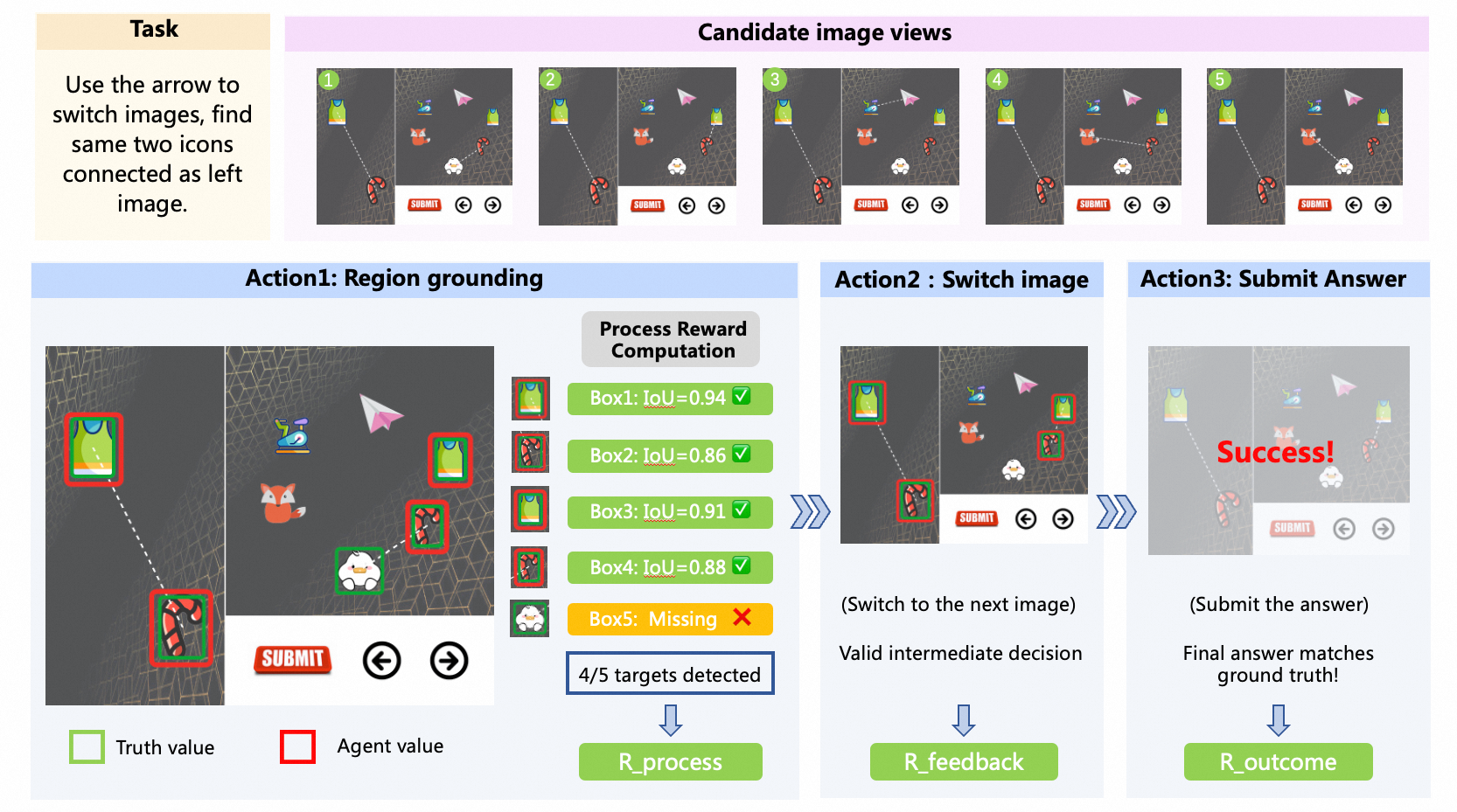}
  \caption{An illustrative training trajectory showing how explicit reasoning steps are supervised at different stages of CAPTCHA solving. The model first performs region grounding to expose task-relevant visual reasoning, receives process-level rewards based on alignment with ground-truth regions, obtains interaction feedback for valid intermediate decisions, and finally receives an outcome reward upon successful task completion.}
  \label{fig:process_supervision}
\end{figure*}

In CAPTCHA environments, the bounding\_box tool produces explicit annotations of key visual regions. Before solving a task, the model indicates which image regions it considers relevant, and these annotations directly expose the model’s intermediate reasoning. 
As illustrated in Figure~\ref{fig:process_supervision}, the model explicitly grounds both the target icons and candidate regions via the bounding\_box tool, thereby exposing its intermediate reasoning about which visual elements are relevant for solving the task. By comparing the predicted regions with the ground-truth regions recorded during data generation, we can directly assess whether the model’s reasoning process is correct and provide supervision accordingly. The bounding\_box tool thus serves a dual role: it functions both as a reasoning aid that helps the model focus on relevant regions and as a supervision interface that makes reasoning quality measurable. This form of process supervision is made possible by our programmatically generated data, in which all key visual elements are precisely recorded.

\paragraph{Design of process-level rewards.}
As shown in Figure~\ref{fig:process_supervision}, we introduce multi-level reward signals that correspond to different stages of the interaction trajectory. 
First, we define a reasoning process reward $R_{\text{process}}$ to directly supervise the model's intermediate reasoning quality. When the model invokes the bounding\_box tool, we evaluate the quality of predicted regions by comparing them with ground-truth key regions. For each predicted bounding box, if its intersection-over-union (IoU) with any ground-truth region exceeds 0.8, we consider it a correct identification. The reasoning process reward is determined by the coverage of correctly identified regions: the model receives a positive reward when it correctly identifies at least 50\% of the ground-truth regions, encouraging sufficient evidence gathering without requiring exhaustive identification of every relevant element. These thresholds are determined through preliminary experiments; sensitivity analysis is provided in Appendix~\ref{sec:sensitivity}.
Second, we introduce an interaction feedback reward $R_{\text{feedback}}$ for tasks requiring multi-step interaction. The environment evaluates whether each action leads to valid progress toward task completion, providing positive feedback when an action is judged to be a correct intermediate step based on the task's state transition logic.
Finally, we use an outcome reward $R_{\text{outcome}}$, which provides positive feedback if the task is successfully completed within an episode and zero otherwise. The overall reward at each step combines these three components:
\begin{equation}
R = R_{\text{process}} + R_{\text{feedback}} + R_{\text{outcome}}.
\end{equation}

\paragraph{Reinforcement learning training.}
During training, we jointly sample episodes from the eight CAPTCHA tasks. The model interacts with the environment over multiple steps, collecting full trajectories consisting of states, actions, and rewards, and we update the policy $\pi_\theta
$ using the GRPO algorithm~\cite{grpo} with 8 rollout samples per prompt. In practice, we selectively apply explicit reasoning process supervision based on task difficulty. For tasks with SFT accuracy below 50\%, the process reward $R_{\text{process}}$ is enabled and plays a crucial role in guiding the model toward correct reasoning behaviors. For tasks where SFT already achieves reasonable performance, we find that training with interaction feedback and outcome rewards alone is sufficient, and thus $R_{\text{process}}$ is set to 0 for these tasks to simplify the reward signal.

\begin{table*}[t]
\centering
\small
\setlength{\tabcolsep}{6pt}
\begin{tabular}{p{3.6cm}c c c c c c c c c}
\hline
\thead[l]{\textbf{Method}} &
\thead[c]{\textbf{Avg.}} &
\thead[c]{connect\\icon} &
\thead[c]{coordi-\\nates} &
\thead[c]{dart\\count} &
\thead[c]{dice\\count} &
\thead[c]{rotation\\match} &
\thead[c]{image\\select} &
\thead[c]{click\\order} &
\thead[c]{slide\\puzzle} \\
\hline

\rowcolor{gray!12}
\multicolumn{10}{l}{\emph{Closed-source VLMs}} \\
Claude-4-Sonnet        & 47.4 & 7.5  & \textbf{93.5} & 68.5 & 51.0 & 20.0 & 43.0 & 71.5 & 24.5 \\
Claude-3.7-Sonnet      & 43.9 & 19.0 & 83.5 & 51.0 & 75.5 & 10.0 & 38.5 & 30.0 & 44.0 \\
Gemini-3-Pro           & 45.8 & 47.5 & 62.0 & 88.5 & 63.0 & 50.0 & 52.0 & 0.0  & 3.0  \\
OpenAI-o3              & 46.9 & 37.0 & 69.0 & 51.0 & 72.0 & 4.0  & 46.0 & 62.0 & 34.0 \\
GPT-5                  & 33.6 & 15.0 & 35.5 & 37.0 & 42.0 & 3.5  & 32.5 & 67.0 & 36.0 \\
Qwen-VL-Max            & 38.6 & 14.5 & 55.5 & 40.0 & 42.5 & 5.5  & 39.5 & 72.5 & 39.0 \\

\hline

\rowcolor{gray!12}
\multicolumn{10}{l}{\emph{Agentic Methods}} \\
Oedipus                & 51.9 & 35.0 & 68.0 & 76.0 & 44.5 & 31.5 & 53.0 & 68.5 & 38.5 \\
Halligan               & 54.7 & 34.0 & 76.5 & 69.5 & 43.0 & 36.0 & 66.0 & 71.0 & 41.5 \\
\hline

\rowcolor{gray!12}
\multicolumn{10}{l}{\emph{GUI Agent}} \\
GUI\_R1\_7b            & 6.5  & 0.0  & 13.5 & 14.0 & 0.5  & 3.5  & 18.0 & 2.0  & 0.0  \\
\hline

\rowcolor{gray!12}
\multicolumn{10}{l}{\emph{Training-based Methods}} \\
SFT-only               & 68.1 & 49.0 & 79.5 & 77.0 & \textbf{76.0} & \textbf{88.0} & 14.0 & 70.5 & 91.0 \\
\textbf{CaptchaMind (Ours)} & \textbf{82.9} & \textbf{71.0} & 91.0 & \textbf{93.0} & 72.5 & 86.0 & \textbf{71.0} & \textbf{87.5} & \textbf{91.5} \\
\hline
\end{tabular}
\caption{Task success rates (\%) of different methods on eight CAPTCHA tasks. Methods are grouped into four categories: closed-source VLMs, agentic methods, GUI agents, and our proposed CaptchaMind. \textbf{Avg.} denotes the average success rate across all eight tasks.}

\label{tab:main_results}
\end{table*}

\section{Experiments}

\subsection{Experimental Setup}

\paragraph{Datasets and training.}
We use Qwen2.5-VL-7B~\cite{qwen25vl} as the base model. The 16,000 generated samples are split into 8,000 for supervised fine-tuning and 8,000 for reinforcement learning, with 1,600 samples (200 per task) reserved for evaluation. Training is performed in two stages on 16 NVIDIA A100 GPUs: supervised fine-tuning followed by reinforcement learning using the GRPO algorithm~\cite{grpo}. Complete training configurations are provided in Appendix~\ref{sec:training_details}. 

\paragraph{Baselines.}
We compare CaptchaMind against the following baselines.
 (1) \emph{Training-based methods}: an SFT-only model (i.e., the model after the first training stage described in Section~\ref{sec:sft}).
 (2) \emph{Closed-source VLMs}: Claude-4-Sonnet~\cite{claude4}, Claude-3.7-Sonnet~\cite{claude37}, Gemini-3-Pro~\cite{gemini3}, Qwen-VL-Max~\cite{qwenvl}, GPT-5~\cite{gpt5}, and OpenAI-o3~\cite{o3}.
 (3) \emph{Agentic methods}: Halligan~\cite{halligan}, a VLM-based agent that formulates CAPTCHA solving as a search problem, and Oedipus~\cite{oedipus}, a solver based on a domain-specific language for CAPTCHA decomposition.
 (4) \emph{GUI agents}: models trained on GUI interaction tasks~\cite{uir1,guir1,infiguir1,osatlas,infiguiagent,uitars2,zerogui}, evaluated to assess whether GUI capabilities transfer to CAPTCHA settings.


\paragraph{Evaluation metric.}
We use task success rate as the primary metric, defined as the proportion of tasks successfully solved within a limited number of steps. All baseline models are evaluated using a unified prompt template detailed in Appendix~\ref{sec:prompts}.

\subsection{Main Results}

Table~\ref{tab:main_results} reports the performance of our method and all baseline approaches. CaptchaMind achieves the highest average success rate of 82.9\%, outperforming all existing methods by a substantial margin.

\paragraph{Limitations of existing methods.}
Existing methods perform poorly on CaptchaBench overall. Closed-source VLMs perform adequately on perceptual tasks but fail substantially on tasks demanding fine-grained visual detail capture (e.g., connect\_icon). Agent-based methods such as Halligan and Oedipus, despite specialized decomposition and search mechanisms, exhibit similar limitations, indicating that prompt-based and agent-based strategies cannot address the core challenges of CAPTCHA solving.

\paragraph{Fine-tuning improves performance but generalizes poorly.}
SFT substantially improves model performance (68.1\% average), but still struggles on certain tasks (e.g., connect\_icon, image\_select), and generalizes poorly to unseen tasks and out-of-distribution data (see Appendix~\ref{sec:generalization_details}).

\begin{figure}[h]
\centering
\begin{tikzpicture}
\begin{axis}[
    width=0.9\columnwidth,
    height=5.5cm,
    ybar,
    bar width=14pt,
    xlabel={Region Identification Rate (\%)},
    ylabel={Task Success Rate (\%)},
    xmin=-0.5, xmax=4.5,
    ymin=0, ymax=100,
    xtick={0,1,2,3,4},
    xticklabels={{[0, 0.2)}, {[0.2, 0.4)}, {[0.4, 0.6)}, {[0.6, 0.8)}, {[0.8, 1.0]}},
    x tick label style={font=\scriptsize, anchor=center},
    y tick label style={font=\scriptsize},
    xtick style={draw=none},
    ytick={0,20,40,60,80,100},
    yticklabel={\pgfmathprintnumber{\tick}\%},
    ymajorgrids=true,
    grid style={dashed, gray!30},
    tick label style={font=\small},
    label style={font=\small},
    nodes near coords,
    nodes near coords align={vertical},
    every node near coord/.append style={font=\tiny},
]
\addplot[fill=blue!55, draw=blue!70] coordinates {
    (0, 16)
    (1, 51)
    (2, 82)
    (3, 90)
    (4, 91)
};
\end{axis}
\end{tikzpicture}
\caption{Task success rate across region identification rate intervals, where region identification rate measures the proportion of ground-truth task-relevant regions correctly identified by CaptchaMind before acting. Higher identification rate consistently leads to higher success rate, confirming that attending to task-relevant visual details is the key determinant of task success.}
\label{fig:correlation}
\end{figure}
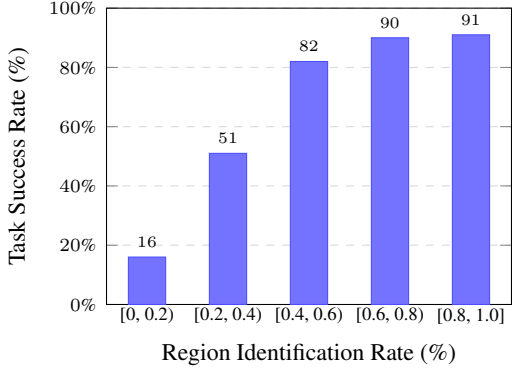

\paragraph{CaptchaMind achieves state-of-the-art performance.}
Our proposed CaptchaMind, trained with explicit reasoning process supervision, improves the model's ability to capture critical visual details, thereby addressing the bottleneck in fine-grained reasoning tasks and achieving 82.9\% average success rate, outperforming all existing methods by a substantial margin. To further validate this, we analyze the relationship between region identification rate and task success rate across all evaluation episodes. As shown in Figure~\ref{fig:correlation}, episodes where fewer than 20\% of task-relevant regions are correctly identified achieve only 16\% success rate, rising to 91\% when more than 60\% are identified, confirming that attending to task-relevant visual details is the key determinant of task success. Experimental details are provided in Appendix~\ref{sec:correlation_analysis}, and a qualitative comparison is provided in Section~\ref{sec:case_study}.

\paragraph{GUI agent skills do not transfer to CAPTCHA tasks.}
Although both involve interface interaction, CAPTCHA tasks additionally demand cross-image reasoning and multi-step decision-making that GUI-oriented training does not cover. Most GUI agents fail to produce valid outputs on CaptchaBench, and the only stably-running model, GUI\_R1\_7b, achieves only 6.5\%, confirming GUI capabilities are insufficient for CAPTCHA solving. Detailed analysis is provided in Appendix~\ref{sec:baseline_failures}.

\subsection{Real-World Evaluation}
\label{sec:sim_to_real}

To comprehensively assess CaptchaMind's generalization beyond CaptchaBench, we evaluate on two additional settings: cross-dataset evaluation on OpenCaptchaWorld (Appendix~\ref{sec:generalization_details}) and real-world evaluation on live CAPTCHA instances, which we detail in this section.

\paragraph{Setup.}
We collect 100 CAPTCHA screenshots from live websites (e.g., GeeTest, hCaptcha) with manually annotated answers, evaluated under the same inference protocol as our main experiments.

\paragraph{Results.}
CaptchaMind achieves 71.0\% success rate on real-world CAPTCHAs, compared to 82.9\% on our CaptchaBench test set. The gap is expected, as real-world instances include task variants and visual styles not present in CaptchaBench. These results demonstrate that CaptchaMind generalizes effectively to real-world CAPTCHAs.

\paragraph{Discussion.}
Together with the human discrimination study in Section~\ref{sec:visual_authenticity}, these results validate the ecological validity of CaptchaBench from complementary perspectives: our synthesized instances are perceptually indistinguishable from real ones, and capabilities learned on CaptchaBench transfer to real-world instances.

\begin{table*}[t]
\centering
\small
\setlength{\tabcolsep}{6pt}
\begin{tabular}{p{2.4cm}ccccccccc}
\hline
\thead[l]{\textbf{Setting}} &
\thead[c]{\textbf{Avg.}} &
\thead[c]{connect\\icon} &
\thead[c]{coordi-\\nates} &
\thead[c]{dart\\count} &
\thead[c]{dice\\count} &
\thead[c]{rotation\\match} &
\thead[c]{image\\select} &
\thead[c]{click\\order} &
\thead[c]{slide\\puzzle} \\
\hline
\textbf{Full} 
& \textbf{82.9} & \textbf{71.0} & \textbf{91.0} & \textbf{93.0} & 72.5 & 86.0 & \textbf{71.0} & \textbf{87.5} & \textbf{91.5} \\

SFT 
& 68.1 & 49.0 & 79.5 & 77.0 & \textbf{76.0} & \textbf{88.0} & 14.0 & 70.5 & 91.0 \\

w/o RPS 
& 78.9 & 56.0 & 83.5 & 89.5 & 76.5 & 89.5 & 54.5 & 90.0 & 92.0 \\

w/o FBR 
& 80.0 & 70.5 & 90.0 & 92.5 & 73.0 & 86.5 & 58.0 & 79.0 & 91.0 \\

RL-only 
& 25.6 & 31.0 & 26.0 & 65.0 & 12.5 & 16.0 & 11.5 & 37.0 & 6.0 \\

\hline
\end{tabular}

\caption{Ablation study results (\%) on our test set. We ablate different components of our training framework. \textbf{Full}: our complete method. \textbf{SFT}: supervised fine-tuning only without reinforcement learning. \textbf{w/o RPS}: RL training without reasoning process supervision. \textbf{w/o FBR}: RL training without feedback reward. \textbf{RL-only}: reinforcement learning without SFT warm-up.}

\label{tab:ablation}
\end{table*}

\begin{figure*}[t]
  \centering
  \includegraphics[width=\textwidth]{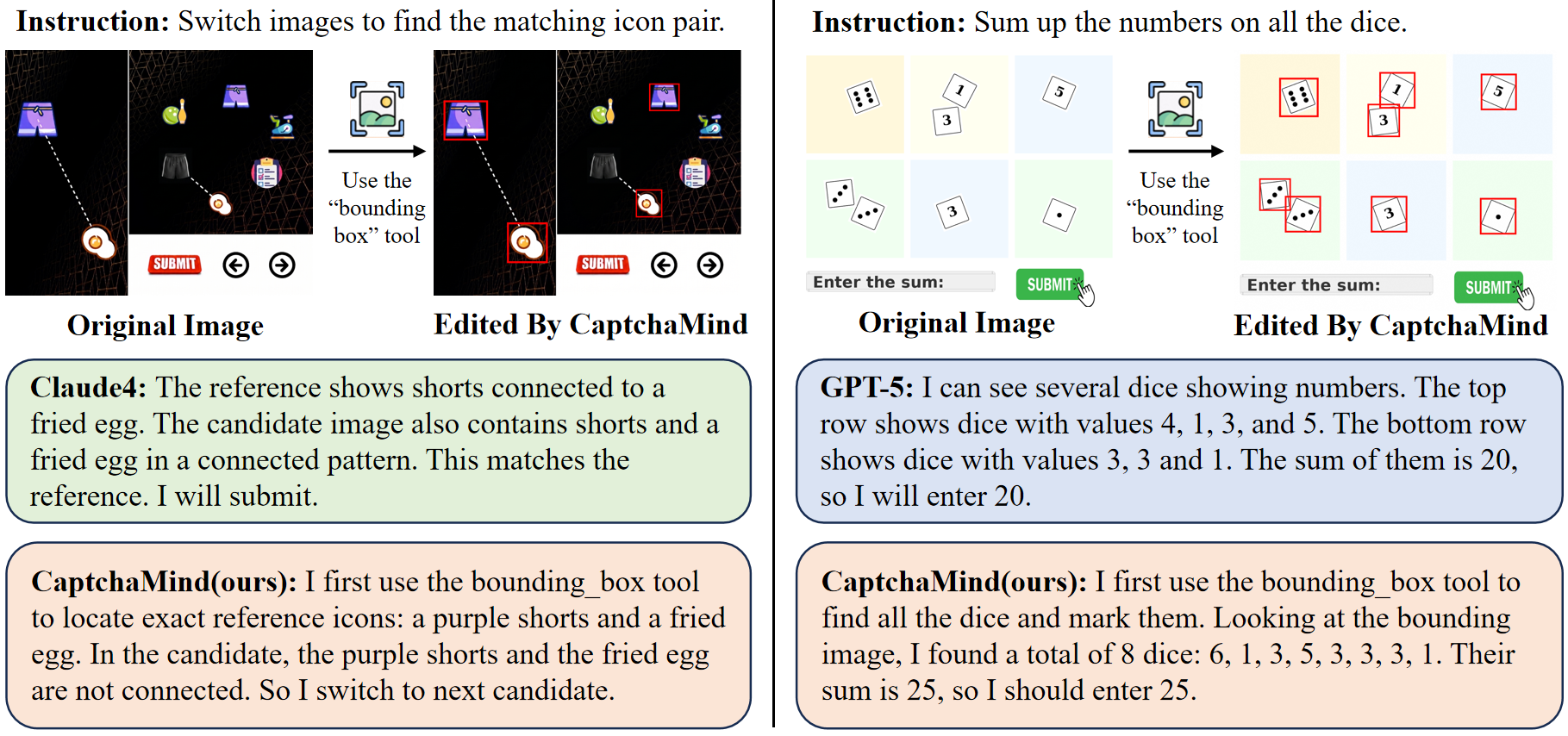}
  \caption{Qualitative comparison on connect\_icon (left) and dice\_count (right). Baseline models make errors by relying on holistic visual impressions, while CaptchaMind consistently attends to task-relevant visual details and makes correct decisions, demonstrating the effectiveness of our training framework.}
  \label{fig:case_analysis}
\end{figure*}

\subsection{Ablation Studies}
\label{sec:ablation}

Table~\ref{tab:ablation} reports the ablation study results. We systematically remove different components of our training framework to assess their contributions.

\paragraph{Reasoning process supervision is critical for challenging tasks.}
We introduce explicit reasoning process supervision for tasks whose SFT accuracy is below 50\%, namely connect\_icon and image\_select --- precisely the tasks requiring fine-grained visual detail capture. Removing $R_{\text{process}}$ leads to substantial performance drops on both tasks, while having minimal impact on simpler tasks where SFT already performs well, confirming that process supervision is most effective on tasks demanding precise region-level attention.

\paragraph{Feedback reward provides guidance for multi-step interaction tasks.}
Removing $R_{\text{feedback}}$ leads to noticeable performance drops on multi-step interaction tasks such as image\_select and click\_order, while having minimal impact on single-step tasks where no intermediate feedback exists. The feedback reward provides step-wise guidance for tasks requiring sequential interactions to reach the correct final state.

\paragraph{SFT warm-up is a prerequisite for reinforcement learning.}
Directly applying RL without SFT results in only 25.6\% average accuracy. Without SFT initialization, the initial policy is too weak to complete tasks reliably, leading to extremely sparse rewards that prevent effective policy optimization. SFT raises task accuracy to above 50\% on most tasks, providing sufficiently dense reward signals for RL to proceed.

\subsection{Case Study}
\label{sec:case_study}

Figure~\ref{fig:case_analysis} presents qualitative comparisons on two representative tasks, illustrating how our training framework enables CaptchaMind to capture critical task-relevant visual details that existing methods overlook.

On the connect\_icon task, the candidate image contains two pairs of connected icons, including both a purple shorts and a black shorts each connected to a fried egg. Claude-4-Sonnet relies on overall visual impression and concludes that the candidate matches the reference since both contain shorts connected to a fried egg, without distinguishing which shorts is connected. CaptchaMind precisely locates the reference icons and identifies that the purple shorts and fried egg are not connected in the candidate, correctly rejecting it.

On the dice\_count task, GPT-5 estimates dice values directly from a holistic view, misreading a 6-dot die as 4 and missing one of two closely stacked dice in the lower-left, resulting in an incorrect sum of 20. CaptchaMind identifies all 8 dice individually, including the stacked ones, and correctly computes the sum as 25.

Both cases illustrate that existing methods fail when tasks demand precise attention to visual details, while CaptchaMind consistently captures the critical details for correct decisions. Complete episode demonstrations of CaptchaMind solving representative tasks are provided in Appendix~\ref{sec:case_study_appendix}.

\section{Conclusion}

We present \textbf{CaptchaBench} and \textbf{CaptchaMind} as a unified solution to the long-standing absence of training-based CAPTCHA solvers. CaptchaBench provides the first large-scale training-oriented benchmark with process-level annotations, and systematic evaluation reveals that existing methods fail consistently on tasks requiring fine-grained visual detail capture. CaptchaMind addresses this by training with explicit reasoning process supervision, achieving state-of-the-art performance across all tasks. Further analysis confirms that region grounding accuracy is the key determinant of task success, directly validating that explicitly supervising visual attention is the source of CaptchaMind's effectiveness. Real-world evaluations further confirm that learned capabilities transfer beyond the synthetic domain. We will release all resources to facilitate future research on multimodal agents in interactive environments.

\clearpage 

\section*{Limitations}

Our real-world evaluation focuses on the visual reasoning component of CAPTCHA solving; real-world deployment additionally involves behavioral verification, dynamic rendering, and anti-bot detection mechanisms that fall outside the scope of this work and remain important directions for future research. Our reward function design, particularly the IoU threshold (0.8) and coverage requirement (50\%), is based on preliminary experiments; although sensitivity analysis (Appendix~\ref{sec:sensitivity}) confirms robustness within reasonable ranges, exhaustive hyperparameter search remains future work. Regarding baseline comparisons, since Oedipus~\cite{oedipus} is not open-sourced, we re-implement its core approach based on the paper description, which may introduce slight differences from the original; for Halligan~\cite{halligan}, necessary adaptations for environment compatibility may similarly introduce minor differences.
\section*{Ethical considerations}

We acknowledge that effective CAPTCHA-solving methods could potentially be misused for automated attacks or malicious activities. However, commercial CAPTCHA-solving services already exist in practice. Our work explores a training-based approach with explicit reasoning supervision, which differs methodologically from existing solutions. 
Training and benchmark evaluation experiments were conducted in controlled simulated environments. Real-world evaluation was performed on publicly accessible demo instances provided by commercial CAPTCHA services solely for research purposes, with no interaction with end-user traffic.
We are committed to responsible disclosure and will engage with relevant stakeholders before releasing any trained models.

\bibliography{custom}

\clearpage
\appendix

\section{Task Interface Examples}
\label{sec:task_examples}

This section provides detailed visual examples of all eight CAPTCHA task types used in our experiments.

\subsection{Image-Switching Tasks}

\begin{figure}[H]
\centering
\begin{subfigure}[t]{0.48\columnwidth}
    \centering
    \includegraphics[width=\textwidth,height=3.5cm,keepaspectratio]{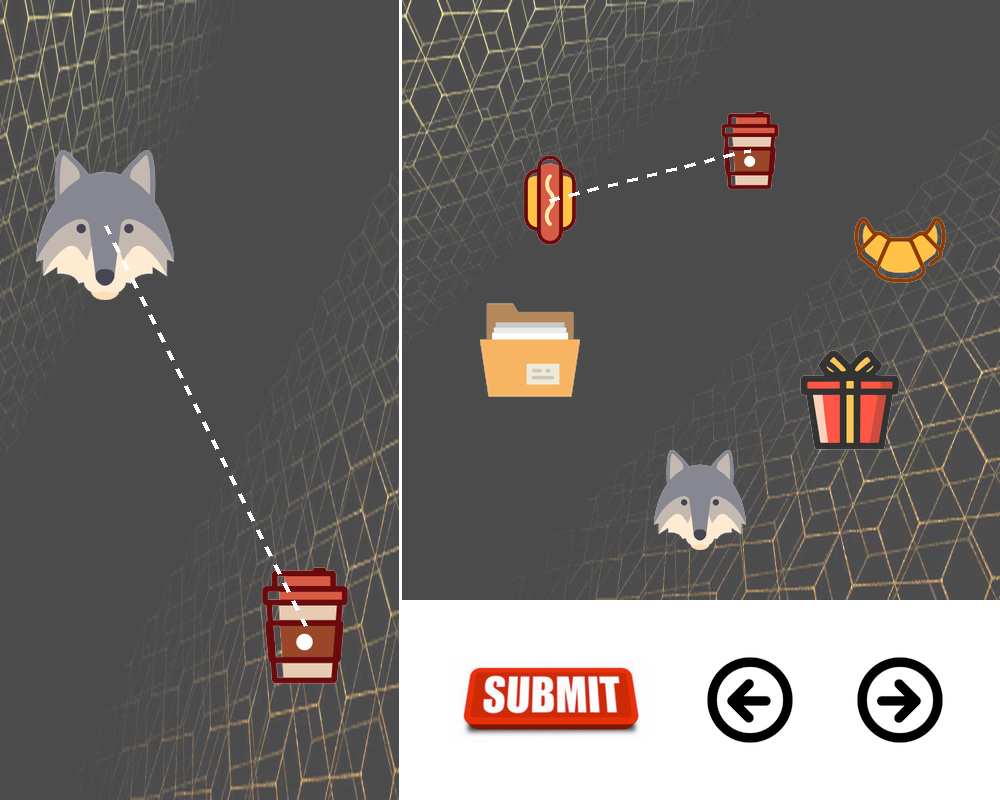}
    \caption{Connect Icon}
    \label{fig:connect_icon}
\end{subfigure}
\hfill
\begin{subfigure}[t]{0.48\columnwidth}
    \centering
    \includegraphics[width=\textwidth,height=3.5cm,keepaspectratio]{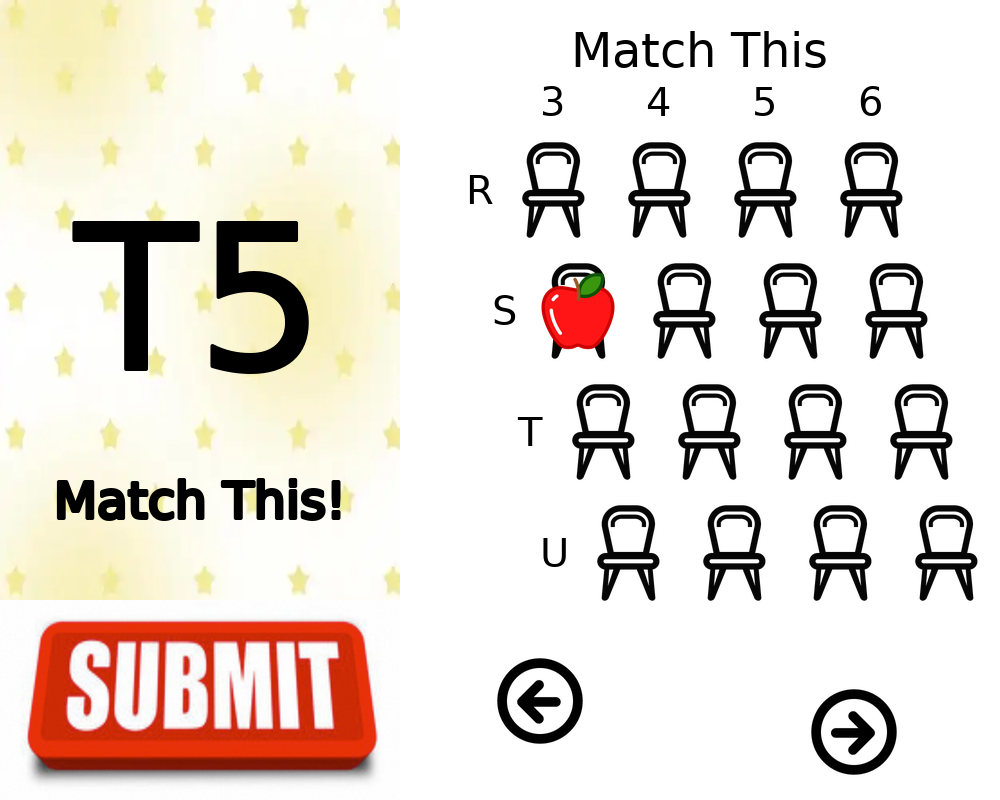}
    \caption{Coordinates}
    \label{fig:coordinates}
\end{subfigure}

\vspace{0.3cm}

\begin{subfigure}[t]{0.48\columnwidth}
    \centering
    \includegraphics[width=\textwidth,height=3.5cm,keepaspectratio]{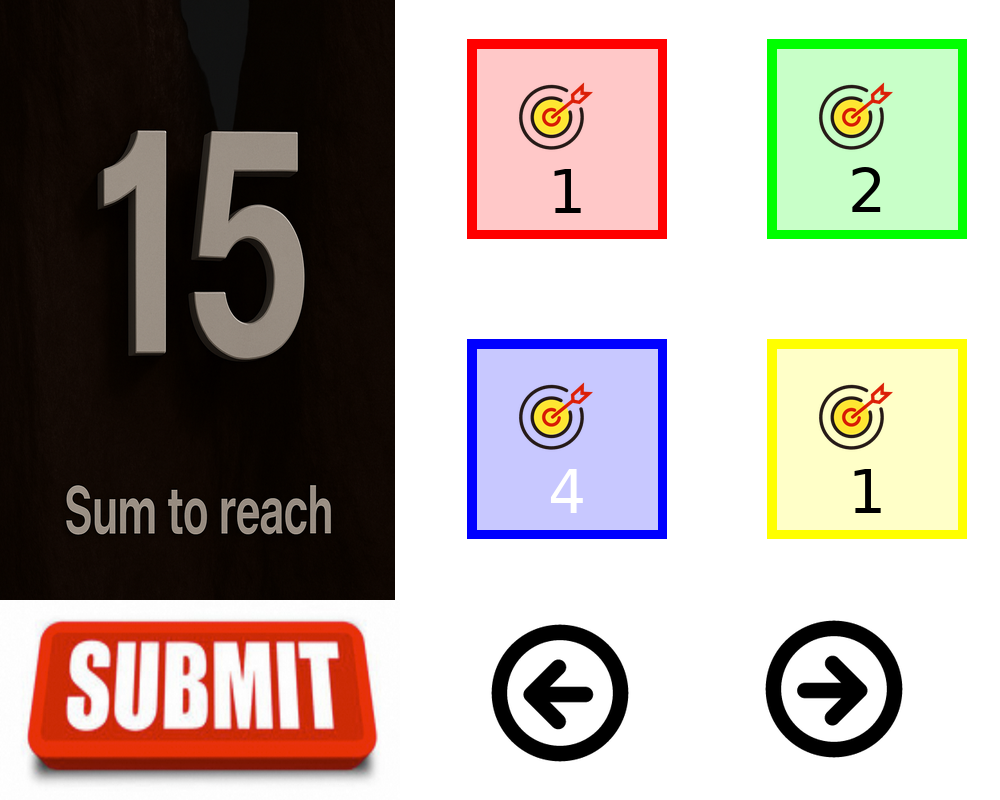}
    \caption{Dart Count}
    \label{fig:dart_count}
\end{subfigure}
\hfill
\begin{subfigure}[t]{0.48\columnwidth}
    \centering
    \includegraphics[width=\textwidth,height=3.5cm,keepaspectratio]{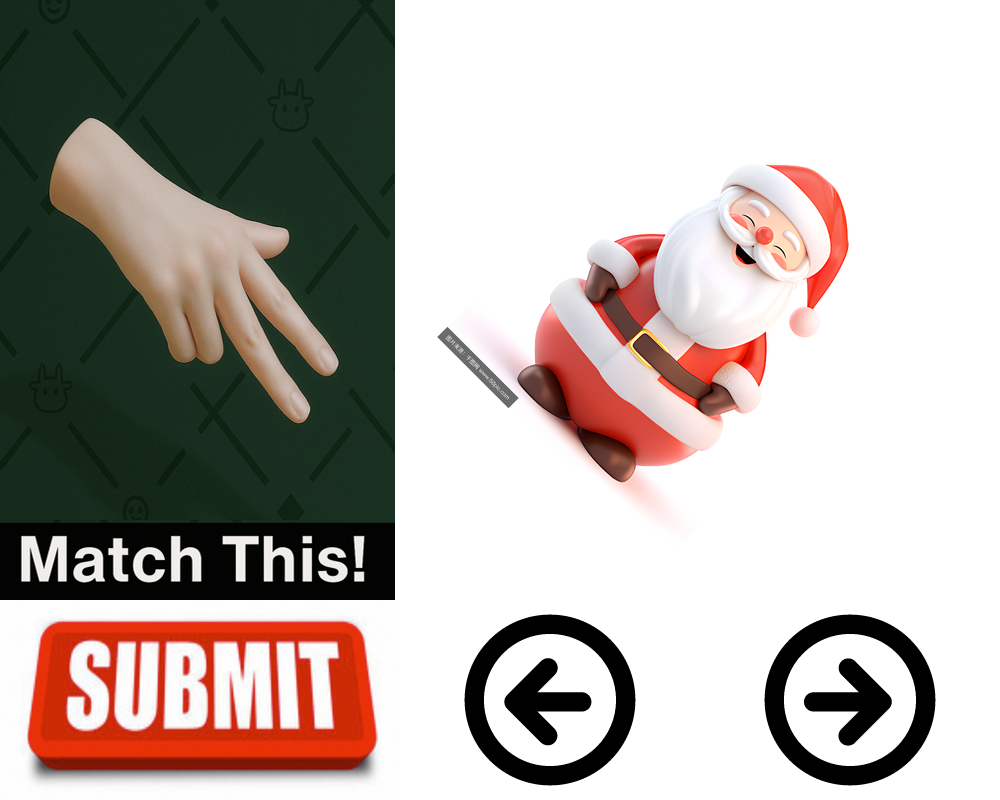}
    \caption{Rotation Match}
    \label{fig:rotation_match}
\end{subfigure}

\caption{Examples of image-switching tasks.}
\label{fig:image_switching}
\end{figure}

As shown in Figure~\ref{fig:image_switching}, in image-switching tasks, the model needs to switch among multiple candidate images and identify the correct option:
\begin{itemize}[leftmargin=*, nosep, after=\vspace{0.5em}]
\item \textbf{Connect Icon:} Use the arrows to switch images, find same two icons connected as left image.
\item \textbf{Coordinates:} Use the arrows to find the image where the animal or fruit is in the correct position.
\item \textbf{Dart Count:} Use the arrows to find the image where all the darts add up to the target number shown on the left.
\item \textbf{Rotation Match:} Use the arrows to find the image where the object faces the same direction as the reference.
\end{itemize}

\subsection{Multi-Step Interaction Tasks}

\begin{figure}[H]
\centering
\begin{subfigure}[t]{0.48\columnwidth}
    \centering
    \includegraphics[width=\textwidth,height=3.5cm,keepaspectratio]{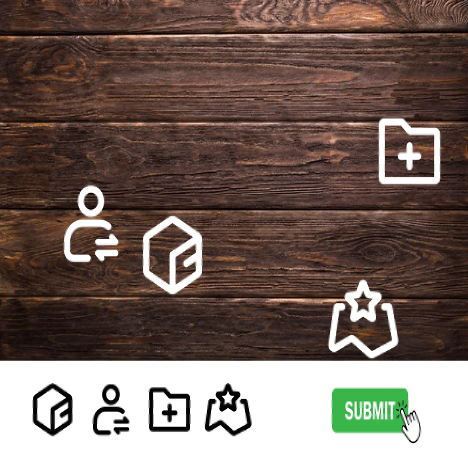}
    \caption{Click Order}
    \label{fig:click_order}
\end{subfigure}
\hfill
\begin{subfigure}[t]{0.48\columnwidth}
    \centering
    \includegraphics[width=\textwidth,height=3.5cm,keepaspectratio]{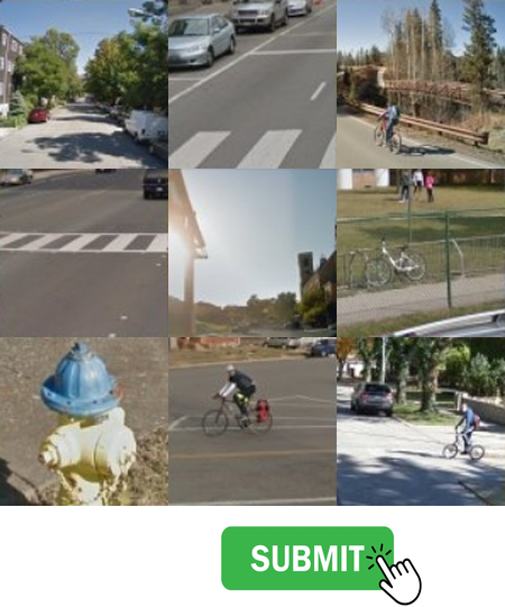}
    \caption{Image Select}
    \label{fig:image_select}
\end{subfigure}

\caption{Examples of multi-step interaction tasks.}
\label{fig:multistep}
\end{figure}

As shown in Figure~\ref{fig:multistep}, in multi-step interaction tasks, the model performs multiple interactions on the same image and receives step-wise feedback:
\begin{itemize}[leftmargin=*, nosep, after=\vspace{0.5em}]
\item \textbf{Click Order:} Click the icons in order as shown in the reference image.
\item \textbf{Image Select:} Select all images containing the bicycle.
\end{itemize}

\subsection{Single-Step Decision Tasks}

\begin{figure}[H]
\centering
\begin{subfigure}[t]{0.48\columnwidth}
    \centering
    \includegraphics[width=\textwidth,height=3.5cm,keepaspectratio]{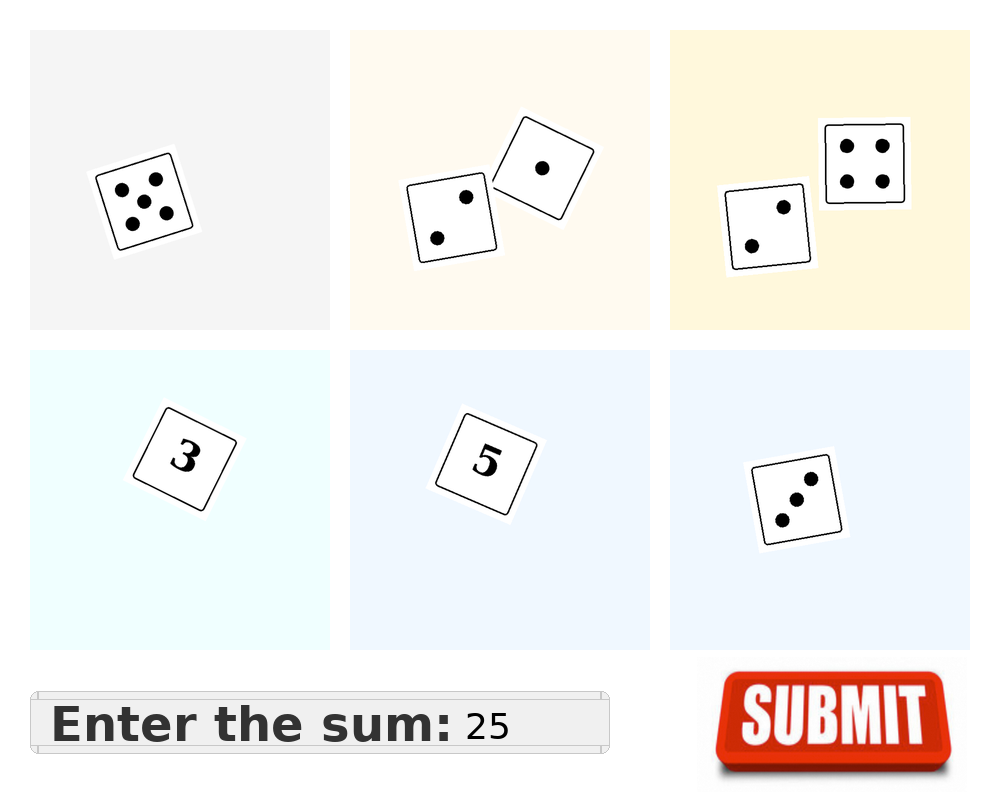}
    \caption{Dice Count}
    \label{fig:dice_count}
\end{subfigure}
\hfill
\begin{subfigure}[t]{0.48\columnwidth}
    \centering
    \includegraphics[width=\textwidth,height=3.5cm,keepaspectratio]{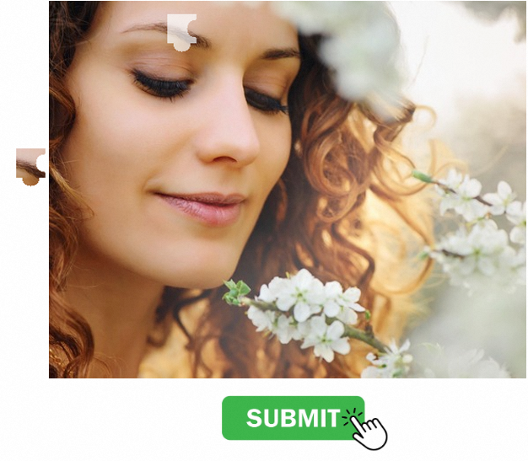}
    \caption{Slide Puzzle}
    \label{fig:slide_puzzle}
\end{subfigure}

\caption{Examples of single-step decision tasks.}
\label{fig:singlestep}
\end{figure}

As shown in Figure~\ref{fig:singlestep}, in single-step decision tasks, the model must produce an accurate result in a single operation:
\begin{itemize}[leftmargin=*, nosep, after=\vspace{0.5em}]
\item \textbf{Dice Count:} Sum up the numbers on all the dice.
\item \textbf{Slide Puzzle:} Drag the slider component to the correct position.
\end{itemize}

\section{Automated Data Generation Details}
\label{sec:data_generation}

This section provides the complete implementation of our automated data generation pipeline.

\subsection{Programmatic Scene Construction and Ground-Truth Recording}

For each task type, we programmatically construct CAPTCHA instances by randomly combining visual elements such as icons, backgrounds, spatial positions, and rotation angles, resulting in samples with diverse appearances and difficulty levels. This randomization includes variations in the number of objects (e.g., 3 to 8 icons), their spatial layout, rotation angles (from -90° to +90° in 15° increments), and the visual styles of backgrounds and elements, ensuring a wide range of difficulty. During data generation, we simultaneously record key ground-truth information, including the regions containing correct answers, task-relevant candidate regions, button locations, interaction feedback, and task success conditions. While such annotations are not directly accessible in real web environments, they can be precisely generated in our simulator, providing the necessary foundation for reasoning process supervision.

\subsection{Multi-Step Interaction Trajectory Generation}

For tasks that require multiple rounds of interaction, we automatically construct complete interaction trajectories, including the correct action sequence and the updated image after each action. Feedback at each interaction step is also recorded, such that each trajectory contains full information in the form of ``state--action--feedback--outcome.'' These trajectories can be used both as training samples for behavior cloning during supervised fine-tuning and as environment transition data during reinforcement learning.

\subsection{Chain-of-Thought Data Generation}

For certain task types, we generate chain-of-thought (CoT) reasoning data using the structured information recorded during scene construction. The reasoning traces are directly derived from ground-truth factors such as true counts and rotation angles, and are expressed in diverse textual forms to avoid inducing dependence on fixed templates during supervised fine-tuning. This process is fully automated and requires no human annotation, allowing it to be naturally extended to new CAPTCHA types.

\section{Training Configuration Details}
\label{sec:training_details}

\subsection{Data Split}

The 16,000 training samples are split as follows:
\begin{itemize}
\item \textbf{Supervised fine-tuning:} 8,000 samples (1,000 per task)
\item \textbf{Reinforcement learning:} 8,000 samples (1,000 per task)
\item \textbf{Evaluation:} 1,600 samples (200 per task)
\end{itemize}

For the leave-one-out experiment, the coordinates task is excluded, resulting in 14,000 training samples.

\subsection{Training Hyperparameters}

Table~\ref{tab:training_config} summarizes the complete training configuration.

\begin{table}[h]
\centering
\small
\begin{tabular}{ll}
\hline
\textbf{Configuration} & \textbf{Value} \\
\hline
\multicolumn{2}{l}{\textit{Model}} \\
Base model & Qwen2.5-VL-7B \\
\hline
\multicolumn{2}{l}{\textit{Hardware}} \\
GPUs & 16 × NVIDIA A100 (80GB) \\
\hline
\multicolumn{2}{l}{\textit{Supervised Fine-Tuning}} \\
Training samples & 8,000 \\
Epochs & 3 \\
Learning rate & 5e-6 \\
\hline
\multicolumn{2}{l}{\textit{Reinforcement Learning}} \\
Training samples & 8,000 \\
Algorithm & GRPO \\
Epochs & 1 \\
Batch size & 64 \\
Rollout samples per prompt & 8 \\
Learning rate & 5e-6 \\
Training time & 20 hours \\
\hline
\end{tabular}
\caption{Complete training configuration and hyperparameters.}
\label{tab:training_config}
\end{table}

\section{Prompts}
\label{sec:prompts}

We use the following prompt template for all CAPTCHA tasks. The \texttt{\{instruction\}} placeholder is replaced with task-specific instructions as described in Appendix~\ref{sec:task_examples}.

\begin{lstlisting}
You are a captcha solver and your task is to solve a captcha step by step. 
{instruction}
At each step, your generation should have exactly the following format:
<think>Your reasoning process to understand the task, analyze the image, and decide what action to take.</think>
<tool_call>{"name": <function-name>, "arguments": <args-json-object>}</tool_call>
Your available actions are bounding, click, drag and enter_number, examples are:
1. {"name": "click", "arguments": {"position": [100, 150]}}
2. {"name": "drag", "arguments": {"from": [100, 150], "to": [100, 200]}}
3. {"name": "bounding", "arguments": {"boxes": [[50, 50, 150, 150], [200, 200, 300, 300]]}} 
4. {"name": "enter_number", "arguments": {"number": 47}}
You should only take an action at a step. After each step, you obtain an observation. Before solving the task, you should first use bounding boxes to mark the key positions. You can click the submit button to submit the final result.
\end{lstlisting}

\section{Generalization Analysis}
\label{sec:generalization_details}

\begin{table*}[h]
\centering
\small
\setlength{\tabcolsep}{6pt}
\begin{tabular}{p{2.4cm}ccccccccc}
\hline
\thead[l]{\textbf{Method}} &
\thead[c]{\textbf{Avg.}} &
\thead[c]{connect\\icon} &
\thead[c]{coordi-\\nates} &
\thead[c]{dart\\count} &
\thead[c]{rotation\\match} &
\thead[c]{image\\select} &
\thead[c]{click\\order} &
\thead[c]{slide\\puzzle} &
\thead[c]{\textbf{Total}} \\
\hline
\textbf{CaptchaMind}
& \textbf{66.2} & 6/10 & 5/10 & 12/20
& 23/40 & 10/20 & 15/20 & 29/31 & \textbf{100/151} \\
SFT-only
& 29.8 & 1/10 & 3/10 & 7/20
& 9/40 & 2/20 & 8/20 & 15/31 & 45/151 \\
Claude-4-Sonnet
& 51.0 & 4/10 & 7/10 & 17/20 & 3/40  & 13/20 & 15/20 & 18/31 & 77/151 \\
Qwen-VL-Max
& 41.7 & 6/10 & 3/10 & 4/20  & 4/40  & 6/20  & 17/20 & 22/31 & 63/151 \\
Gemini-3-Pro
& 23.8 & 4/10 & 4/10 & 8/20  & 12/40 & 7/20  & 0/20  & 1/31  & 36/151 \\
\hline
\end{tabular}
\caption{Cross-dataset evaluation results on the OpenCaptchaWorld benchmark. We include the best-performing closed-source VLMs from our main experiments as baselines. SFT-only generalizes substantially worse than CaptchaMind, demonstrating that RL with explicit reasoning process supervision learns more transferable reasoning abilities.}
\label{tab:open_captcha_world}
\end{table*}

To evaluate robustness under distribution shifts, we evaluate on the OpenCaptchaWorld~\cite{opencaptchaworld} benchmark, covering 151 samples across task types that overlap with our training set but differ in data sources and visual styles. We include the best-performing closed-source VLMs from our main experiments as additional baselines. As shown in Table~\ref{tab:open_captcha_world}, the SFT-only model achieves only 29.8\%, lower than most closed-source baselines, confirming that SFT alone generalizes poorly to out-of-distribution data. CaptchaMind achieves 66.2\%, outperforming all baselines by a substantial margin despite the distribution shift, demonstrating that CaptchaMind's capabilities generalize effectively across different data distributions.

\section{Reward Threshold Sensitivity Analysis}
\label{sec:sensitivity}

Our reward function relies on two thresholds: the IoU threshold for bounding box matching and the coverage threshold for region identification. Table~\ref{tab:sensitivity} reports model performance across different combinations of these two thresholds.

\begin{table}[h]
\centering
\small
\setlength{\tabcolsep}{5pt}
\begin{tabular}{ccccccc}
\hline
\multirow{2}{*}{\textbf{IoU}} & \multicolumn{6}{c}{\textbf{Coverage Threshold}} \\
& 30\% & 40\% & 50\% & 60\% & 70\% & 80\% \\
\hline
0.6 & 80.9 & 82.4 & 82.9 & 82.8 & 82.2 & 81.3 \\
0.8 & 80.8 & 82.7 & 82.9 & 83.1 & 82.3 & 81.4 \\
0.9 & 80.6 & 82.8 & 82.7 & 82.4 & 82.5 & 81.1 \\
\hline
\end{tabular}
\caption{Sensitivity analysis of reward thresholds (\%). Rows denote IoU thresholds; columns denote coverage thresholds. Our default setting (IoU=0.8, Coverage=50\%) is highlighted in bold.}
\label{tab:sensitivity}
\end{table}

For the IoU threshold, we observe that the model either localizes regions precisely or misses them entirely, making the exact threshold non-critical. Results confirm this: performance remains stable across a wide range of IoU thresholds (0.6--0.9). For coverage, performance peaks around 50\%--60\% and degrades slightly at higher thresholds (e.g., 80\%), where the model is overly penalized for efficient reasoning paths that identify only the most task-relevant regions. These results confirm that our method is robust to the specific choice of reward thresholds within reasonable ranges.


\section{Region Grounding Accuracy and Task Success Correlation}
\label{sec:correlation_analysis}

To directly validate that capturing task-relevant visual details drives task success, we analyze the relationship between region grounding coverage and task success rate across all 1,600 evaluation episodes of CaptchaMind.

\paragraph{Metric.}
Following the process reward definition in Section~\ref{sec:rl_supervision}, we compute a grounding coverage score for each episode. A predicted bounding box is considered correct if its IoU with any ground-truth region exceeds 0.8. The grounding coverage score is defined as the ratio of correctly identified regions to the total number of annotated ground-truth regions recorded during data generation.

\paragraph{Results.}
As shown in Figure~\ref{fig:correlation}, task success rate increases substantially with grounding coverage. Episodes with coverage below 0.2 achieve only 16\% success rate, rising to 51\% for coverage in [0.2, 0.4), and 82\% for coverage in [0.4, 0.6). Performance further saturates above 0.6, reaching 90\%--91\%. These results provide direct evidence that the ability to identify task-relevant visual regions at intermediate reasoning steps is the key determinant of task success in CAPTCHA solving.

\section{GUI Agent Failure Analysis}
\label{sec:baseline_failures}

\begin{figure}[h]
\centering
\includegraphics[width=\columnwidth]{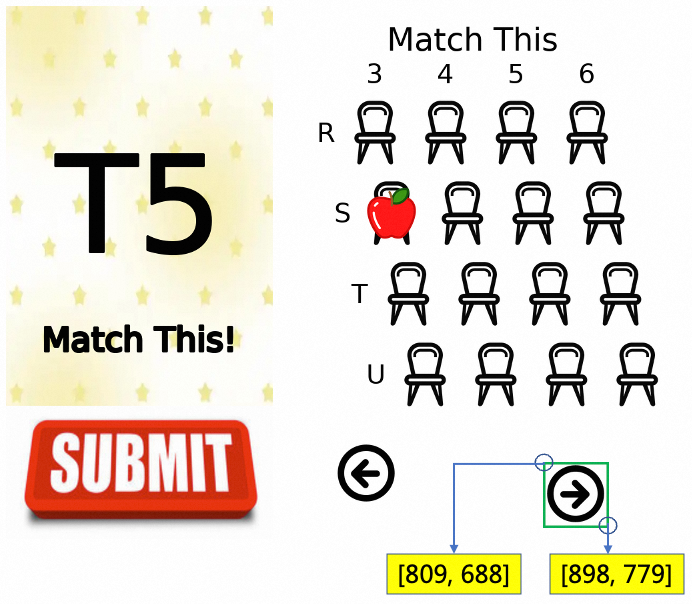}
\caption{A coordinates task where the right arrow button's valid clickable region is bounded by [809, 688] (top-left) and [898, 779] (bottom-right).}
\label{fig:gpt4o_failure}
\end{figure}

GUI agents exhibit two fundamental limitations on CAPTCHA tasks: output format incompatibility and lack of multi-step reasoning capabilities.

\paragraph{Output Format Incompatibility.}
Many GUI agents produce outputs in fixed formats incompatible with CAPTCHA environments. UI-Tars-7B~\cite{uitars2}, for example, generates HTML table structures that lack the required structured format and provide no extractable coordinate information. Although we attempted to modify our parsing logic to accommodate various output formats, most models still failed to produce evaluable results.

\paragraph{Lack of Multi-Step Reasoning.}
Even GUI agents that produce compatible output formats fail due to insufficient reasoning capabilities. OS-Atlas-Base-7B~\cite{osatlas} can generate valid tool calls, but its interaction pattern reveals shallow reasoning. As shown in Figure~\ref{fig:gpt4o_failure}, on a coordinates task, it produces:
\begin{lstlisting}
<think>I see a right arrow button. I will click it.</think>
<tool_call>{"name": "click", "arguments": {"position": [850, 730]}}</tool_call>
\end{lstlisting}
While the coordinates correctly fall within the button region, the model shows no understanding of \emph{why} it should click --- there is no analysis of whether the current image matches the target. It simply clicks any visible button, reflecting the nature of GUI training which primarily consists of single-step element localization tasks requiring minimal reasoning.

\section{Case Study: CaptchaMind Reasoning Process}
\label{sec:case_study_appendix}

\begin{figure*}[h]
\centering
\includegraphics[width=\textwidth]{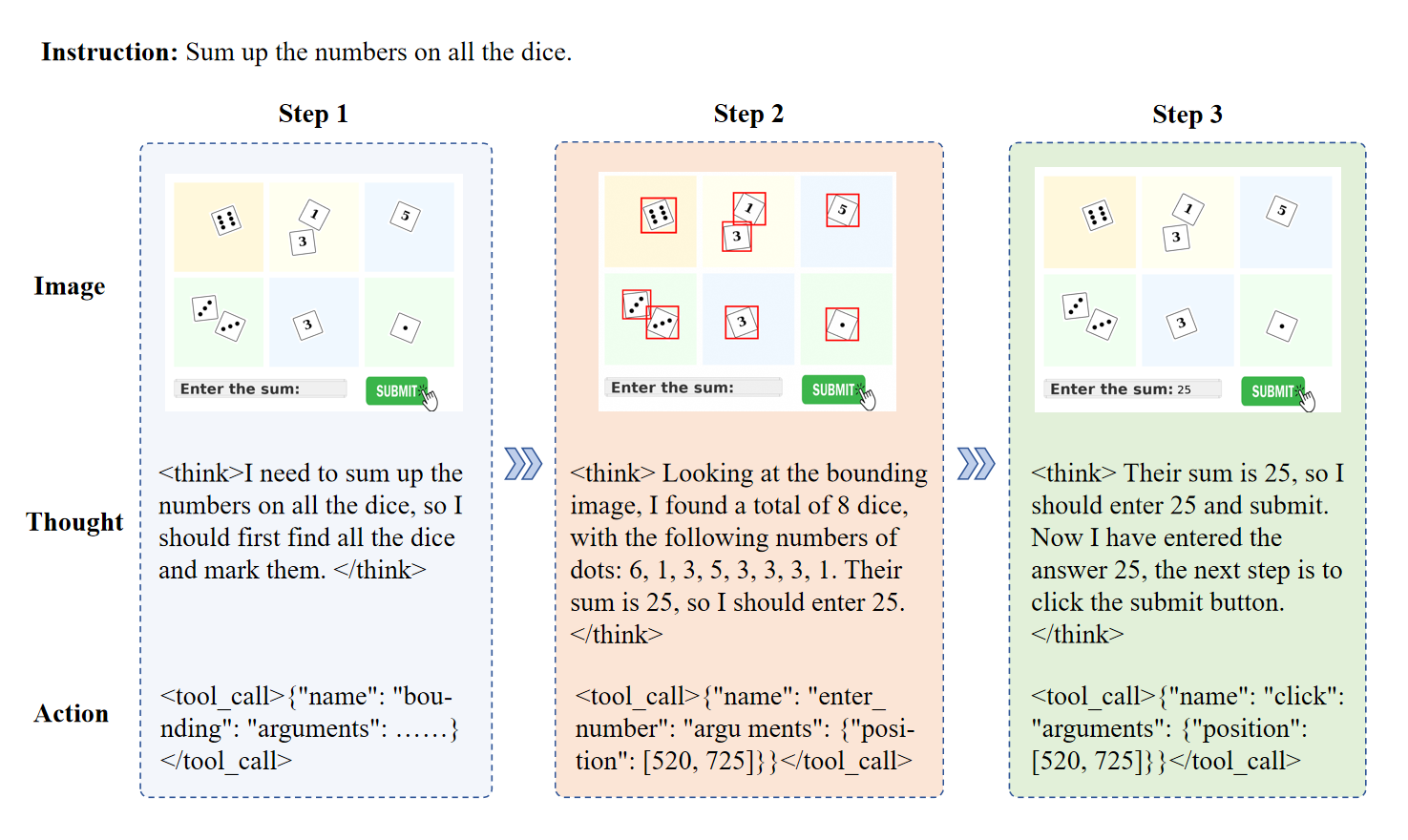}
\caption{CaptchaMind on the \textbf{dice\_count} task. CaptchaMind first uses the bounding box tool to locate all dice (Step 1), reasons over the annotated image to compute the sum (Step 2), and submits the correct answer (Step 3).}
\label{fig:case_dice}
\end{figure*}

\begin{figure*}[h]
\centering
\includegraphics[width=\textwidth]{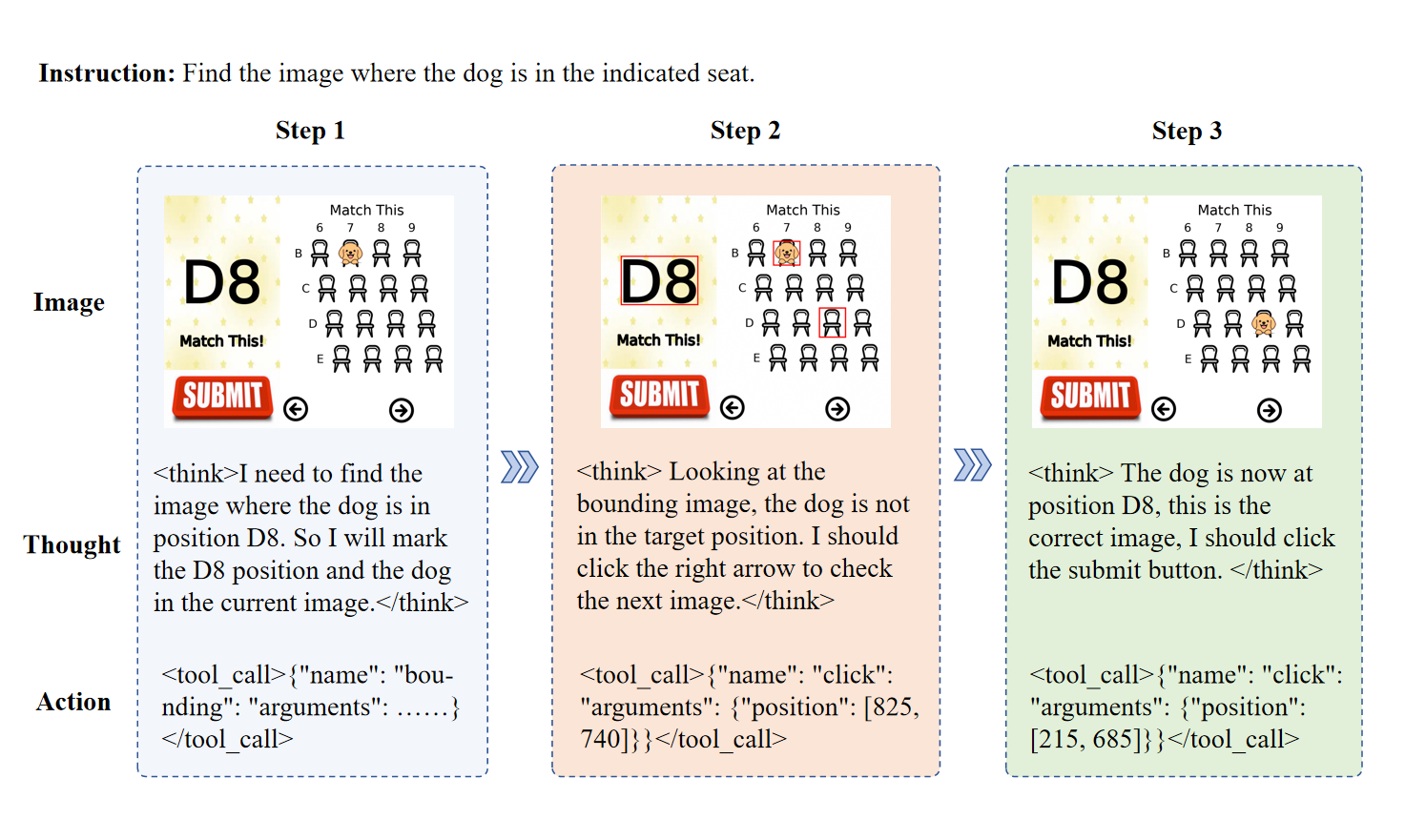}
\caption{CaptchaMind on the \textbf{coordinates} task. CaptchaMind marks the target position and object in the current image (Step 1), identifies a mismatch and switches candidate (Step 2), and submits upon finding the correct match (Step 3).}
\label{fig:case_coord}
\end{figure*}

Figure~\ref{fig:case_dice} and Figure~\ref{fig:case_coord} illustrate representative success cases of CaptchaMind on two task types, demonstrating its complete step-by-step reasoning process across different task types.

\section{Human Study Interface}
\label{sec:study_interface}

\begin{figure*}[h]
\centering
\includegraphics[width=0.9\textwidth]{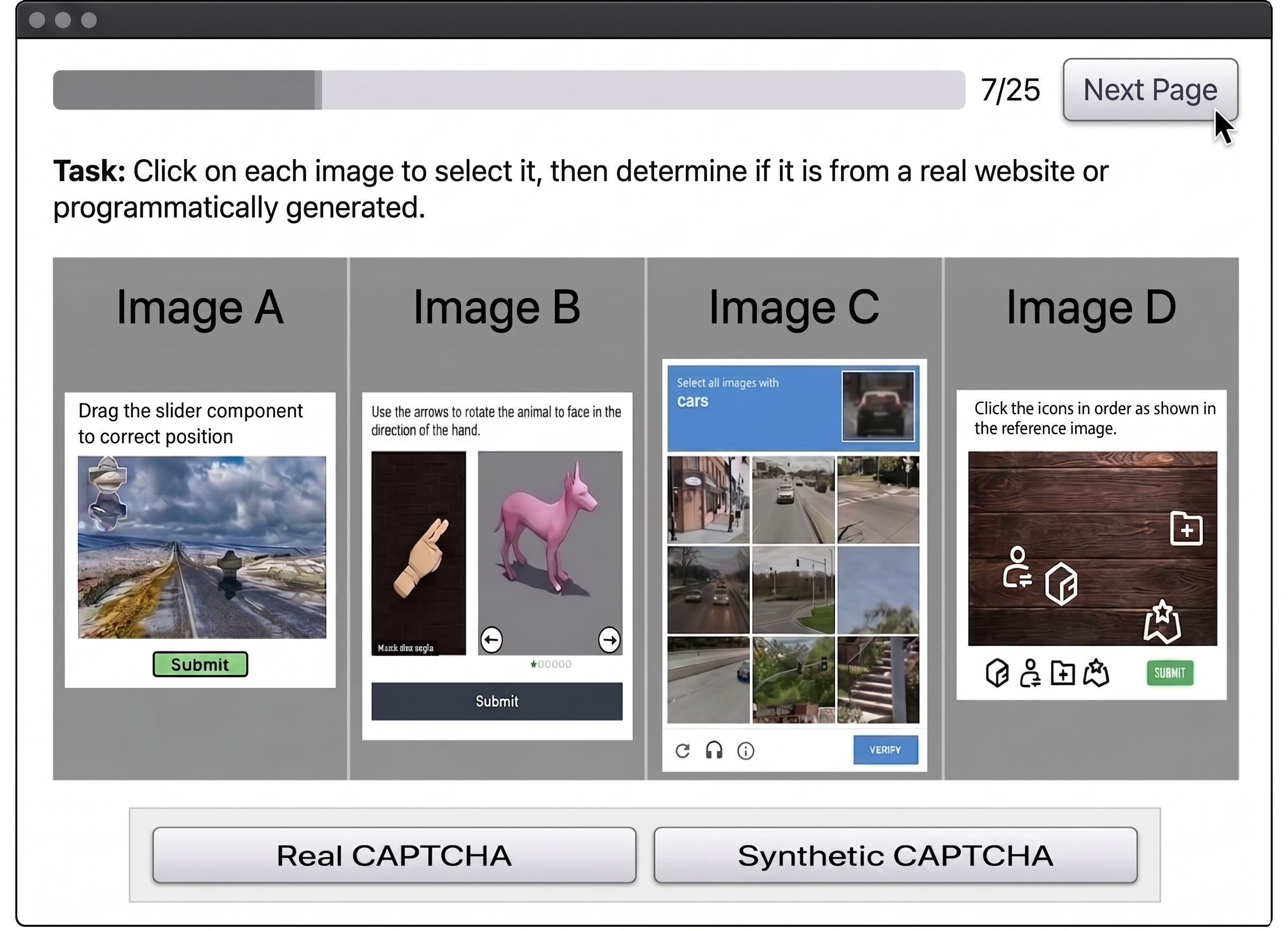}
\caption{The web-based interface used in the human discrimination study.}
\label{fig:study_interface}
\end{figure*}

Figure~\ref{fig:study_interface} shows the web-based interface used in our human discrimination study. Participants were presented with four CAPTCHA images per page and asked to classify each as ``Real CAPTCHA'' or ``Synthetic CAPTCHA.''

\end{document}